\documentclass[letterpaper, 10 pt, conference]{ieeeconf}  

\IEEEoverridecommandlockouts                              

\usepackage{graphics} 
\usepackage{epsfig} 
\usepackage{amsmath} 
\usepackage{amssymb}  
\usepackage{booktabs}
\usepackage{multirow}
\usepackage{multicol}
\usepackage[ruled,vlined]{algorithm2e}

\renewcommand{\vec}[1]{\mathbf{#1}}
\newcommand{\tabincell}[2]{\begin{tabular}{@{}#1@{}}#2\end{tabular}}

\title{\LARGE \bf
ColCOS$\Phi$: A Multiple Pheromone Communication System for Swarm Robotics and Social Insects Research
}

\author{Xuelong Sun$^{1,+}$, Tian Liu$^{1,2,+}$, Cheng Hu$^{2,1}$, Qinbing Fu$^{2,1}$ and Shigang Yue$^{2,1,*}$
\thanks{This work was supported by EU Horizon 2020 project STEP2DYNA (691154) and ULTRACEPT (778062).}
\thanks{
$^{1}$ Computational Intelligence Lab (CIL), School of Computer Science, University of Lincoln, Lincoln, UK, LN6 7TS.}
\thanks{
${2}$ School of Mechanical and Electrical Engineering, Guangzhou University, China, 510006.
}
\thanks{
$^+$ These authors contributed equally to this research
}
\thanks{
$^{*}$ Corresponding author: Shigang Yue. {\tt\small syue@lincoln.ac.uk}}
}

\begin{document}

\maketitle
\thispagestyle{empty}
\pagestyle{empty}

\begin{abstract}
In the last few decades we have witnessed how the pheromone of social insect has become a rich inspiration source of swarm robotics. By utilising the virtual pheromone in physical swarm robot system to coordinate individuals and realise direct/indirect inter-robot communications like the social insect, stigmergic behaviour has emerged. However, many studies only take one single pheromone into account in solving swarm problems, which is not the case in real insects. In the real social insect world, diverse behaviours, complex collective performances and flexible transition from one state to another are guided by different kinds of pheromones and their interactions. Therefore, whether multiple pheromone based strategy can inspire swarm robotics research, and inversely how the performances of swarm robots controlled by multiple pheromones bring inspirations to explain the social insects' behaviours will become an interesting question. Thus, to provide a reliable system to undertake the multiple pheromone study, in this paper, we specifically proposed and realised a multiple pheromone communication system called ColCOS$\Phi$. This system consists of a virtual pheromone sub-system wherein the multiple pheromone is represented by a colour image displayed on a screen, and the micro-robots platform designed for swarm robotics applications. Two case studies are undertaken to verify the effectiveness of this system: one is the multiple pheromone based on an ant's forage and another is the interactions of aggregation and alarm pheromones. The experimental results demonstrate the feasibility of ColCOS$\Phi$ and its great potential in directing swarm robotics and social insects research.
\end{abstract}

\section{INTRODUCTION}
One of the mysterious substances that endow social insects the great ability to collectively solve complex problems and complete giant projects is the pheromone \cite{Camazine2003,Garnier2007}. Such a compound secreted by insects enables the direct and indirect communications between individuals, leading to the emergence of stigmergic behaviours, for example, the trail pheromone of the ants \cite{Holldobler1990} and the alarm pheromone of the honeybees \cite{Robinson1987}. Inspired by social insects' abilities to utilise pheromones to coordinate and cooperate with each other, the pheromone based communications have been studied and applied in swarm robots \cite{Brambilla2013}. Among these studies, the most popular topic is the ant's trail pheromone for foraging and path searching \cite{Sugawara2004,Fujisawa2013}. There are also some other applications and studies of pheromones, such as using repellent pheromone to disperse robots in an large area \cite{Pearce2006}, to distribute robots to carry out a rescue operation \cite{Silva2010}, to emplace the robots to cover an unknown environment \cite{Payton2001} and to implement the robots' aggregation behaviours \cite{Arvin2018}. 

However, all the aforementioned studies only take one single pheromone into account for solving a specific task, which is inconsistent with the situation where real insects usually apply cue interactions of multiple pheromones. For example, aside from the attractive trail pheromone released by the ant during food recruitment, another repellent pheromone is also released \cite{Robinson2005} and a further study demonstrated that at least three useful pheromones are utilised in Pharaoh's ant's (\emph{Monomorium pharaonis}) food recruitment \cite{Robinson2008a}. 
Another fact supporting the multiple pheromone's function is that in the insect society, there are many types of pheromones working together to play the vital role in the species survival and reproduction \cite{Blum1972}. 
According to these phenomena, it appears that the multiple-pheromones decide many collective and complex tasks of social insects instead of a single-pheromone.
However, in the swarm robot domain, little attention has been paid to the scenario wherein multiple pheromones are involved to set up the communication between multi-robots and then influence the behaviour. Thus, more efforts and attentions should be paid to the multiple pheromone research.
Outcomes of this kind of research will not only provide new solutions for swarm robot problems, but may bring inspiration to biologists who are interested in pheromones of social insects. In order to efficiently and conveniently undertake multiple pheromone research, a multiple pheromone communication system is needed.

Although some attempts have been made to let real robots deposit physical substance in the real environment \cite{Russell1997,Purnamadjaja2007,Fujisawa2013}, it is impractical for a low-cost robot to leave physical or chemical substance in the real environment, so here comes the term \emph{virtual pheromone}, which means that pheromone is emulated or represented by other substance in the experiment setting. Since the first attempt of implementing the virtual pheromone \cite{Payton2001}, many studies have tried different methods to implement it. Table \ref{tb:ComVirPhero} compares the most popular methods for doing this. From the comparison, we can see that using optical ways can best emulate all the characteristics of the pheromone (evaporation, diffusion, combination and diversity) and the screen is more robust than the projector for the ambient light's impact. 

\begin{table*}[]
\centering
\caption{Comparison of implementations of virtual pheromone}
\label{tb:ComVirPhero}
\resizebox{\textwidth}{!}{%
\begin{tabular}{@{}ccccc@{}}
\toprule
Substance & Method & \begin{tabular}[c]{@{}c@{}}Corresponding \\ Sensor\end{tabular} & \begin{tabular}[c]{@{}c@{}}Multiple\\pheromone?\end{tabular} & Extra Description \\ \midrule
\multirow{4}{*}{Optics} & \begin{tabular}[c]{@{}c@{}}Localisation system \\ and projector \cite{Garnier2007,Sugawara2004,Simonin2011}\end{tabular} & \multirow{4}{*}{\begin{tabular}[c]{@{}c@{}}Colour sensor /\\ colour camera\end{tabular}} & Yes. Colour encoding & \begin{tabular}[c]{@{}c@{}}Controllable. Flexible to modify parameters of \\ evaporation, diffusion.\\ \textbf{Unstable to the ambient light.}\end{tabular} \\ \cmidrule(lr){2-2} \cmidrule(l){4-5} 
 & \begin{tabular}[c]{@{}c@{}}Localisation system \\ and screen \cite{Arvin2015}\end{tabular} &  & Yes. Colour encoding & \begin{tabular}[c]{@{}c@{}}\textbf{Controllable, stable, flexible to modify parameters of} \\ \textbf{evaporation, diffusion.}\end{tabular} \\ \cmidrule(lr){2-2} \cmidrule(l){4-5} 
 & Pen and touch screen \cite{Kitamura2009} &  & Yes. Color encoding & \begin{tabular}[c]{@{}c@{}}Controllable, stable, flexible to modify parameters of \\ evaporation, diffusion \\ \textbf{Limited number of robots and unable to identify different robots}\end{tabular}\\\cmidrule(lr){2-2} \cmidrule(l){4-5} 
 & \begin{tabular}[c]{@{}c@{}}UVLED and phosphorescent \\ glow-paint \cite{Mayet2010}\end{tabular} &  & \textbf{No.} & Not very controllable and hard to modify the parameters. \\ \midrule
Heat & heater and paraffin wax \cite{Russell1997}& temperature sensor & \textbf{No.} & \textbf{Impractical for small robots, energy-consuming.} \\ \midrule
\multirow{3}{*}{\begin{tabular}[c]{@{}c@{}}Data \\ information\end{tabular}} & IR communication \cite{Payton2001} & \multirow{3}{*}{\begin{tabular}[c]{@{}c@{}}data sender \\ and receiver\end{tabular}} & Yes. Specific data encoding & Cannot implement all the properties of pheromone \\ \cmidrule(lr){2-2} \cmidrule(l){4-5} 
 & \begin{tabular}[c]{@{}c@{}}Localisation based virtual \\ environment \cite{Susnea2009,Reina2017,Valentini2018}\end{tabular}&  & Yes. & \begin{tabular}[c]{@{}c@{}}Pheromone information is calculated and stored \\ in a central computer. \\ \textbf{No direct interactions with robots.}\end{tabular} \\ \cmidrule(lr){2-2} \cmidrule(l){4-5} 
 & RFID tags \cite{Mamei2005,Sakakibara2007,Khaliq2015}&  & Yes. Specific data encoding & Very different to implement all the properties of pheromone \\ \midrule
Chemical substances & ethanol \cite{Fujisawa2013}, eucalyptus oil \cite{Purnamadjaja2007}& gas sensor & Yes. Using different substances.& \textbf{Not very controllable, impractical for micro-robots.} \\ \bottomrule
\end{tabular}%
}
\end{table*}

Therefore, here we propose a multiple pheromone communication system named ColCOS$\Phi$ wherein the pheromone is emulated in an optical way and displayed on a screen. To verify the effectiveness and demonstrate its potential in swarm robotics and social insects research, we will present two case studies, the first is the robot simulation of the multiple pheromone based on an ant's foraging and the second is the interaction of aggregation and the alarm pheromone. Experimental results of both case studies strongly demonstrate the effectiveness and flexibility of this system. Potential applications of this system in the future can be predicted given the significance and value of research concerning multiple pheromone based swarm robotics and multiple pheromone interactions of social insects.

The other sections completing this paper are organised as follows: Section 2 describes the details of the components in this proposed ColCOS$\Phi$ system including the modelling of the virtual pheromone, Section 3 introduces the methods of pheromone field construction and the robot control for the two case studies, Section 4 shows and discusses some results of these two case studies and Section 5 gives the conclusion and future work.

\section{Multiple Pheromone Communication}

\begin{figure*}
    \centering
    \includegraphics[scale=0.48]{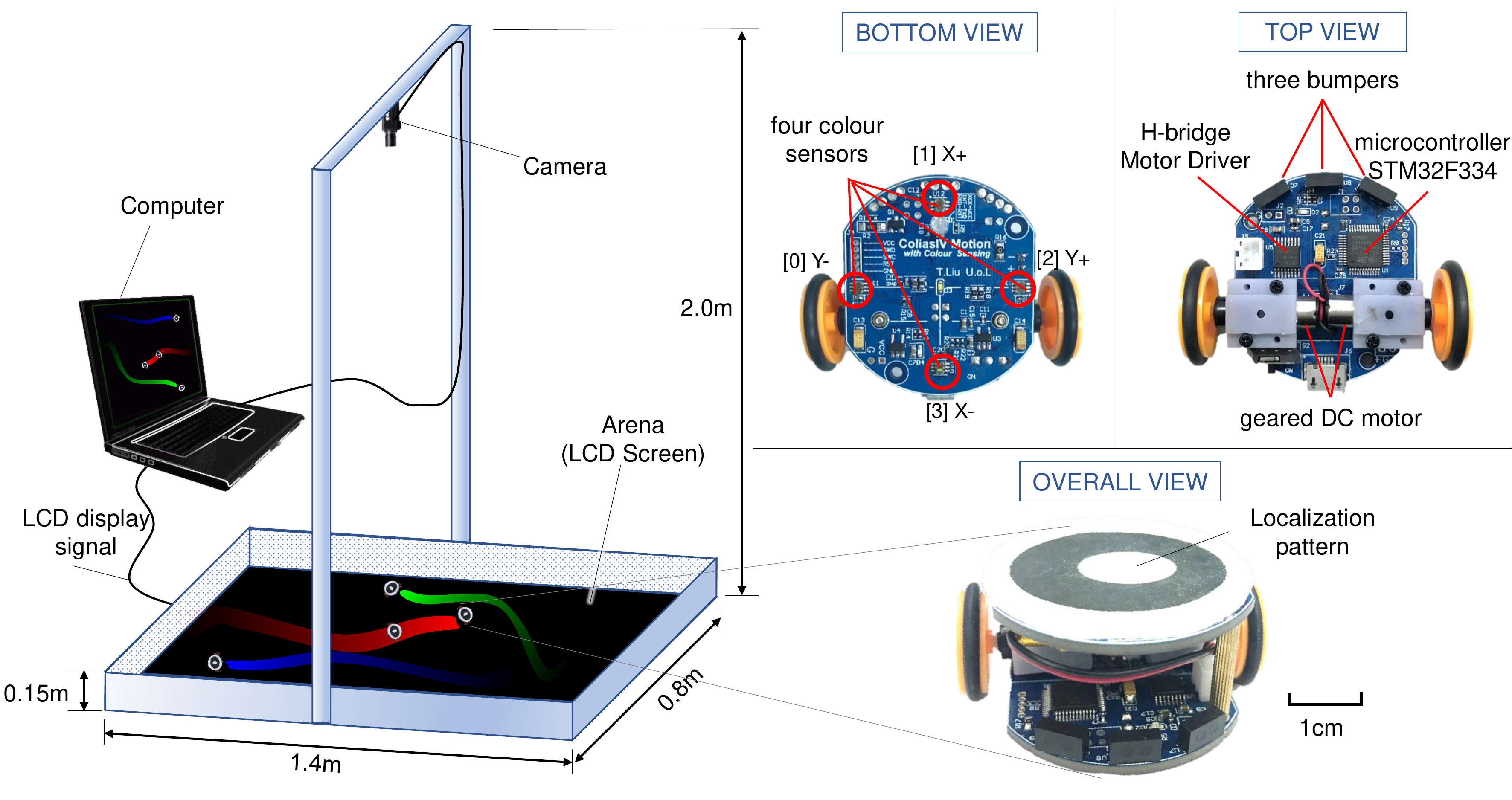}
    \caption{The ColCOS$\Phi$ system. The camera overhead recieves the image containing the patterns attached on the robots, the visual tracking algorithm running in the computer takes this image as input and then outputs the positions of the robots, that is the localisation system. The pheromone field (represented by a colour image) is calculated in the computer by \eqref{eq:PheroDET} or \eqref{eq:PheroGaussian} and updated on the LCD screen every 20ms, with the protecting wall settlling around the screen, it becomes the arena in which the micro-robot explore. Three different views of the micro-robot are showed in the right-hand image.}
    \label{fig:ColCOSP}
\end{figure*}

ColCOS$\Phi$ is a flexible and low-cost virtual pheromone communication system with the great ability to dynamically and precisely simulate multiple pheromones via the three primary colours. This easy-to-use system is an ideal platform that studies swarm robotics and the pheromone interaction of social insects. ColCOS$\Phi$ (Fig. \ref{fig:ColCOSP}) consists of three parts: 
a) a localisation system visually tracking the real-time positions of multi-robots each with an ID-specific pattern,
b) a colour LCD screen which the multiple pheromones are released upon and which the robots explore and exploit, and 
c) a micro robot platform designed specifically for detecting the pheromones on the screen using colour sensors. 
In this section we will introduce this system from the following four aspects: how the pheromone is modelled in this system, the screen that acts as the arena, the localisation system and the robot platform.

\subsection{Pheromone Model}
The virtual pheromone system can simultaneously emulate different kinds of pheromones released by different robots. The whole pheromone field is displayed on the screen as a colour image. 
Therefore, this system can simulate in total three independent types of pheromone encoded by the three primary colours, i.e., red, green and blue.
A colour image can be defined by a $W \times H \times 3$ matrix $\vec{I}$, where $W$ and $H$ is the width and height of the image. Each pixel's value at the position $(x,y)$ of $c$ channel at a specific time $t$ is denoted as $\vec{I}(x,y,c)$, where $x\in N^+$ and $0\le x < W$, $y \in N^+$ and $0\le y < H$ and $c\in\{r,g,b\}$. In our artificial pheromone system, the strength of different pheromones collectively determine the matrix $\vec{I}$ of the colour image displayed on the screen as the linear dynamic system as showed in equation \eqref{eq:ColorImgMatrix}:
\begin{equation}
    \vec{I}(x,y,c) = \sum_{i=0}^{N}{k_{i,j}\vec{\Phi}_{i,j}(x,y)}
    \label{eq:ColorImgMatrix}
\end{equation}
where $\vec{\Phi}_{i,j}(x,y)$ is a $W \times H$ matrix defining the strength of $i^{th}$ type of $j^{th}$ pheromone at location $(x,y)$ and note that the colour channel of the image $c$ is corresponding to the type of the pheromone $j$. $k_{i,j}$ is the effect factor which determine the influence of the pheromone on the image to be displayed on the screen.

To model the important dynamic properties of the pheromone, mainly the evaporation and the diffusion, here we provide two ways to achieve this. First, the differential equation which is showed in \eqref{eq:PheroDET}:

\begin{equation}
    \dot{\vec{\Phi}}_{i,j}(x,y) = -\frac{1}{e_{\Phi_{i,j}}}\vec{\Phi}_{i,j}(x,y) + d_{i,j}\Delta\vec{\Phi}_{i,j}(x,y) + J_{i,j}(x,y)
    \label{eq:PheroDET}
\end{equation}

where:
\begin{itemize}
    \item $e_{\Phi_{i,j}}$ is the evaporation factor which defines how quickly the strength of the pheromone decays exponentially over time.
    \item $d_{i,j}$ is the diffusion factor determining the spread speed of the pheromone. $\Delta\vec{\Phi}_{i,j}(x,y)$ is the spatial rate of the strength change of the pheromone $\vec{\Phi}_{i,j}$, which is calculated by \eqref{eq:PheroDES1} and \eqref{eq:PheroDES2}, where $h$ and $v$ indicate the horizontal and vertical direction respectively.
    \item $J_{i,j}(x,y)$ is the pheromone injection at position $(x,y)$. In the realistic situation, $(x,y)$ can be the robots' current positions or the position where the source of the pheromone designed to be deposited at.
\end{itemize}

\begin{equation}
 \Delta\vec{\Phi}_{i,j}(x,y) = \Delta\vec{\Phi}_{i,j}(x,y)_h + \Delta\vec{\Phi}_{i,j}(x,y)_v
\label{eq:PheroDES1}
\end{equation}

\begin{equation}
\left\{
\begin{array}{lr}
 \Delta\vec{\Phi}_{i,j}(x,y)_h = (\vec{\Phi}_{i,j}(x-1,y) - \vec{\Phi}_{i,j}(x,y))/2 &\\
 \Delta\vec{\Phi}_{i,j}(x,y)_v = (\vec{\Phi}_{i,j}(x,y-1) - \vec{\Phi}_{i,j}(x,y))/2 &
\end{array}
\right.
\label{eq:PheroDES2}
\end{equation}

The dynamic properties of the pheromone are controlled by the above parameters temporally and spatially, which embody all the important characteristics of a real pheromone. 

Another way to model the pheromone in a spatiotemporal way is to apply the scaled bivariate normal distribution and the exponential decay function as showed in \eqref{eq:PheroGaussian}, 
\begin{equation}
\begin{split}
\vec{\Phi}(x,y,t) = &\frac{K}{2\pi\sigma_x\sigma_y\sqrt{1-\rho^2}}\cdot e^{-\frac{t}{e_{\Phi}}}\\
&e^{-\frac{1}{2\left(1-\rho^2\right)}\left[\frac{\left(x-\mu_x\right)^2}{\sigma_x^2}+\frac{\left(y-\mu_y\right)^2}{\sigma_y^2}-\frac{2\rho\left(x-\mu_x\right)\left(y-\mu_y\right)}{\sigma_x\sigma_y}\right]}    
\end{split}
\label{eq:PheroGaussian}
\end{equation}
where:
\begin{itemize}
    \item $K$ is the scale factor determining the strengthen of the pheromone.
    \item $e_{\Phi}$ is the evaporation factor which defines how quickly the strength of pheromone decays exponentially over time $t$.
    \item $\sigma_x$ and $\sigma_y$ is the diffusion factor determining the spread speed of the pheromone in $x$ and $y$ directions respectively. $\rho$ validates the basic different spatial diffusion in the $x$ and $y$ directions, which can be used, in some circumstances, to model the wind's effect on the diffusion.
    \item $\mu_x$ and $\mu_y$ define the position at which the pheromone is injected, which has the same physical meaning with $J_{i,j}(x,y)$ in \eqref{eq:PheroDET}.
\end{itemize}

These two models can be applied according to different 
requirements. Differential equation \eqref{eq:PheroDET} has better dynamic properties and acts more like a real pheromone while with the Gaussian distribution \eqref{eq:PheroGaussian} it is easier to speculate the shape of the pheromone distribution and to get a stable pheromone state.

\subsection{Arena}
In order to precisely display the virtual pheromone, we selected a large and high resolution (4K) colour screen as the arena for the system. It is a 65 inch colour LCD screen placed horizontally which is shown in the left side of Fig. \ref{fig:ColCOSP}. The colour image of the emulated pheromone is produced and refreshed every 20ms by a PC according to the pheromone model introduced in the last section. the robots will explore and interact on the screen. 
In addition, a frame made from a foam PVC sheet is settled around the screen like a wall to protect the robot from running out of the arena. See Fig. \ref{fig:ColCOSP} for the details.

\subsection{Localisation}
The localisation system is based on a fast and precise visual tracking algorithm \cite{Krajnik2013} \cite{Krajnik2014} which implements multi-robots localisation in real time. The hardware that makes up this system is a standard PC and a camera. The core part of this visual tracking method is a novel and efficient algorithm that could recognise white-and-black patterns. A typical pattern is shown in the right part of Fig. \ref{fig:ColCOSP}. This system claims to track hundreds of patterns in real-time using a standard PC \cite{Krajnik2014}. In the follow-up research work, Farshad et al. \cite{Arvin2015} redesigned the patterns and upgraded the software so that the localisation system could not only grasp robots' position $\vec{P}_{i}$, where $i\in \{1,2,3,...,N\} $ and $N$ is the number of robots to be tracked, but also identify individual robot.
The data of robots' position could be regarded as $J_{i,j}$ in \eqref{eq:PheroDET} or ($\mu_x, \mu_y$) in \eqref{eq:PheroGaussian} to calculate the pheromone field.


\subsection{Robot Platform}
On the one hand, the key demand for the robot platform is that the robot could sense the strength of the optic-implemented pheromone. On the other hand, the robot should be small and light so that the arena can accommodate sufficient number of robot for swarm experiments.
In order to factor those requirements, we redesigned the Basic Unit of Colias \cite{Hu2018}. Four colour sensors (TCS34725) are mounted on the bottom of the robot. When the robots are running on the screen, the sensors can directly collect the colour information from the screen which is showed in Fig. \ref{fig:ColCOSP}. In addition, Colias is an affordable platform, and it is specially designed for swarm robotics research with a very compact size of 4cm diameter and is lightweight at approximately 50g. 
To easily establish a Cartesian coordinate system, the distribution of those colour sensors is a $2 \times 2$ array with a 3cm diagonal distance. The sensors are labelled with $n\in{\{0,1,2,3\}}$ corresponding to robot's coordinates ${Y-,X+,Y+,X-}$, which is illustrated in Fig. \ref{fig:ColCOSP}. $\Phi_{i,n}$ stands for the strength of the $i^{th}$ type pheromone. And the robot's pheromone strength $\Phi_i$ define as the average strength of four sensors. Furthermore, the egocentric positive gradient direction can be easily calculated by \eqref{Eq_Gradient}, which also benefits from the specific arrangement of the four sensors.

\begin{equation}
\theta_{i} = \arctan{\frac{\Phi_{i,2}-\Phi_{i,0}}{\Phi_{i,1}-\Phi_{i,3}}}
\label{Eq_Gradient}
\end{equation}

\begin{figure}
    \centering
    \includegraphics[scale=0.29]{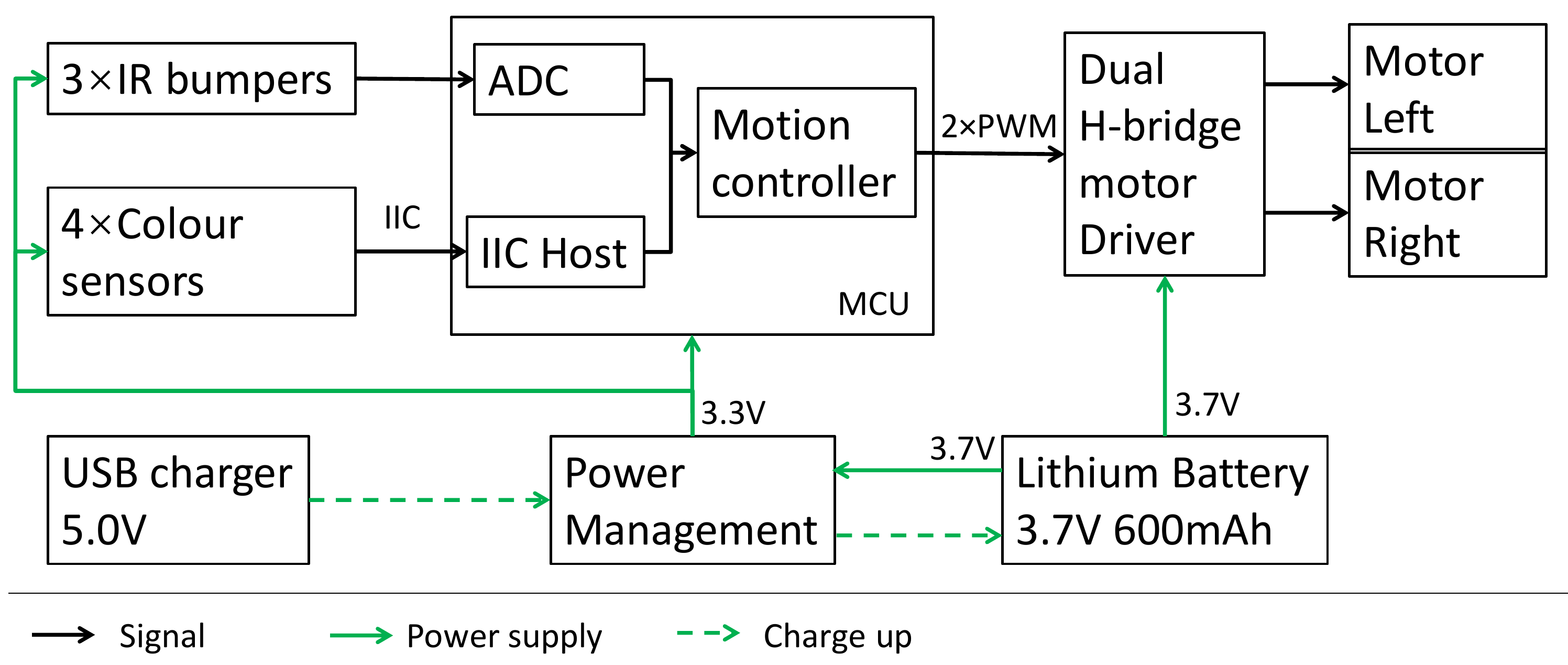}
    \caption{The schematic diagram of the hardware connections of the robot.}
    \label{fig:ArchitectureOfRobot}
\end{figure}

Fig. \ref{fig:ArchitectureOfRobot} shows the schematic diagram of the hardware connections of the robot platform. STM32F334 is chosen as the micro-controller unit (MCU) based on its high-performance ARM-32bit-Cortex-M4 CPU, operating at a frequency of up to 72 MHz. On the basis of its powerful computing power, we used 8 GPIOs for simulating four IIC interfaces to collect the data of four colour sensors. This chip also provides two fast 12-bit analogue-to-digital converters (ADCs), which are used for converting the analogue signal of three bumper sensors into digital data. When the robot is approaching obstacles, the bumper sensor will give a signal that notifies the robot to avoid the collision within a short distance of about 2cm. 
The motion could be controlled by the rotational speeds of the left and the right wheels. As shown in Fig. \ref{fig:ColCOSP}, the two wheels are connected with two geared DC motors. Those two motors are individually driven by two pulse-width modulation (PWM) signals via dual H-bridge motor driver. The micro controller collects the sensory data and then transfer it to the motion command according to the programmed algorithm. (examples can be found in our case studies). 


 
  

\section{Case Study}
In order to verify the effectiveness and availability of ColCOS$\Phi$, we undertake two case studies: one is inspired by the ant's foraging task where at least three pheromones are involved in  \cite{Robinson2008a}. Another one is inspired by the widely existing aggregation and alarm pheromone in social insects.

\subsection{Case I : Three Pheromones in Ant's Foraging}
\subsubsection{Pheromone Field Construction}
\begin{figure}
    \centering
    \includegraphics[scale=0.22]{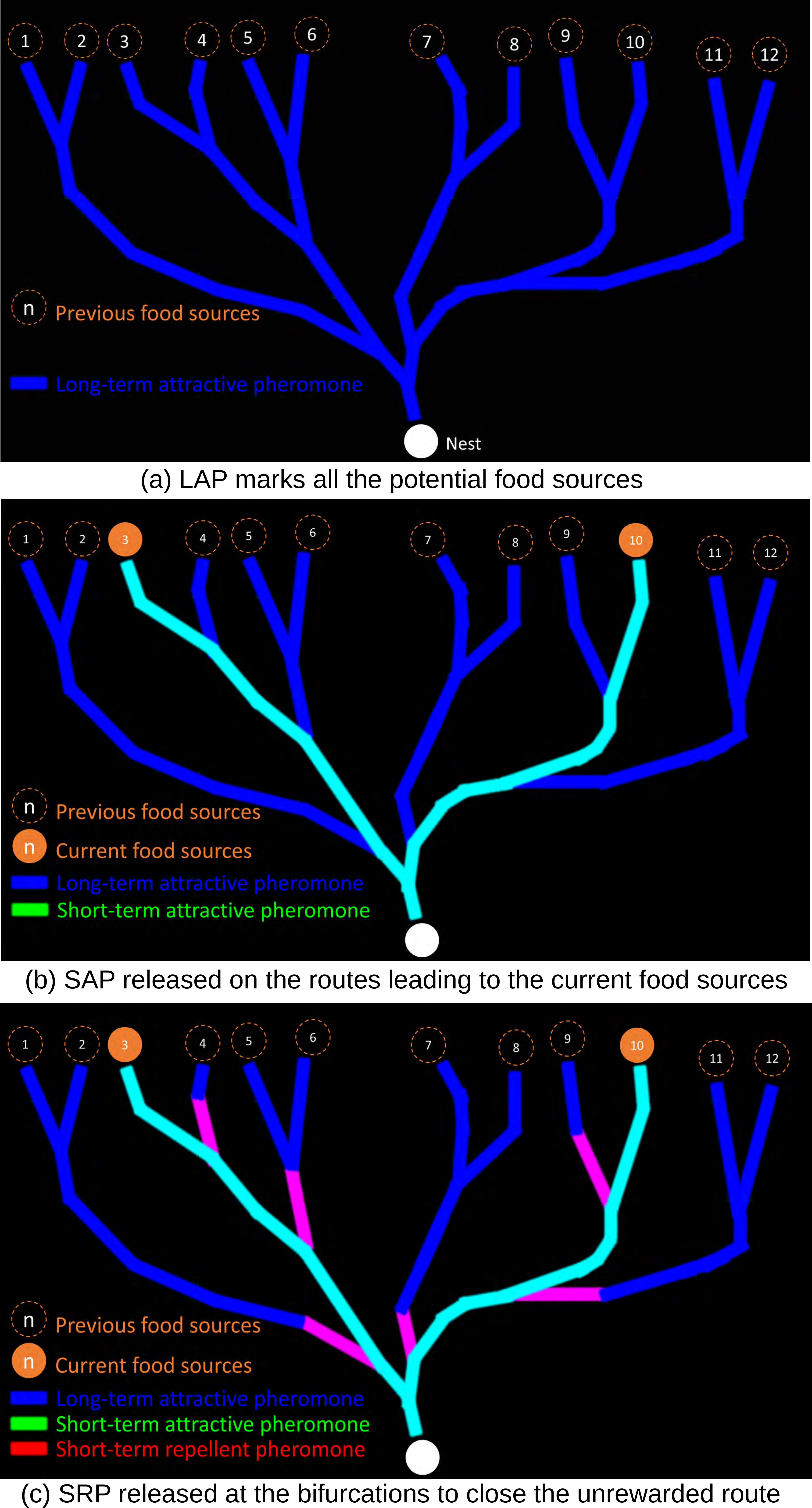}
    \caption{The three kinds of pheromones involved in the ant's foraging task. Three pheromones are indicated by red, green and blue respectively, which mixed together into cyan and magenta.}
    \label{fig:AntForageTrial}
\end{figure}

It is reported that three different types of pheromones are released by the ants which format the environment in which the ants are exploring for food.  They are:
\begin{itemize}
    \item Long-term attractive pheromone (LAP) which constructs the way from the nest to the potential food resources within the environment. This type of pheromone is depicted in Fig. \ref{fig:AntForageTrial} as the blue trails.
    \item Short-term attractive pheromone (SAP) released to convey the up-to-date information of the present available food sources. This pheromone is drawn with green colour that mixes with the blue into cyan in Fig. \ref{fig:AntForageTrial}.
    \item Short-term repellent pheromone (SRP) which is released at the bifurcations to close off the unrewarding branches. This pheromone is drawn with red colour and mixes with the blue into magenta in Fig. \ref{fig:AntForageTrial}.
\end{itemize}

In order to grasp the key point of how different types of pheromones work together to improve the  foraging efficiency, we make an assumption that the pheromone branches have been constructed already and will not change during the same group of experiment. The pheromone used in this case is modelled by \eqref{eq:PheroDET} with very slow evaporation and no diffusion (other parameters used in this case can be found in Table \ref{tb:CS1_PP}). The pheromone map in this case is presented by $\vec{D1},\vec{D2},\vec{D3}$ which contains all the values of the pheromone injections $J_{i,j}(x,y), j=1,2,3$.

\begin{table}
\renewcommand\arraystretch{2}         
\caption{Pheromone Parameter Setting in Case study I}
\centering 
\begin{tabular}{cccccc}
\toprule
Type\hspace{-5pt}&Color\hspace{-5pt}&Model\hspace{-5pt}& Evaporation\hspace{-5pt}&Diffusion\hspace{-5pt}&Injection\hspace{-5pt}\\
\midrule
LAP\hspace{-5pt}&Blue\hspace{-10pt}&\eqref{eq:PheroDET}& \hspace{-5pt}$e_{\Phi_{i,3}} = 50s$&$d_{i,3} = 0$&$J_{i,3} = \vec{D3}_{i}$\\
SAP\hspace{-5pt}&Green\hspace{-10pt}&\eqref{eq:PheroDET}& \hspace{-5pt}$e_{\Phi_{i,2}} = 50s$&$d_{i,2} = 0$&$J_{i,2} = \vec{D2}_{i}$\\
SRP\hspace{-5pt}&Red\hspace{-10pt}&\eqref{eq:PheroDET}& \hspace{-5pt}$e_{\Phi_{i,1}} = 50s$&$d_{i,1} = 0$&$J_{i,1} = \vec{D1}_{i}$\\
\bottomrule
\end{tabular}
\label{tb:CS1_PP}
\end{table}

\subsubsection{Robot Control}
The aim of the agent in this experiment is to get the food resource as correct as possible according to the pheromones deposited in the environment. 
The algorithm designed to control the robot to get this aim is described in Algorithm \ref{algo:AntForaging}. 
It's worth mentioning that in Algorithm \ref{algo:AntForaging} the cyan trail is mixed by blue and green pheromones, similarly, the magenta trail is consisted of blue and red pheromones. The avoiding motion is implemented as a sharp turn from the current moving direction. And the method of controlling the robot to follow a trail is ruled by \eqref{eq:Case1Control}:

\begin{equation}
\left\{
\begin{array}{lr}
    R_l = (\Phi_{l,i} - \Phi_{r,i}) * p + v_b &\\
    R_r = (\Phi_{r,i} - \Phi_{l,i}) * p + v_b
\end{array}
\right.
    \label{eq:Case1Control}
\end{equation}

where:
\begin{itemize}
    \item $R_l$ and $R_r$ is the rotational speed of left wheel and right wheel respectively.
    \item $\Phi_{l,i}$ and $\Phi_{r,i}$ is the left and the right colour sensor's strength of $i^{th}$ type of pheromone respectively.
    \item $p$ is a sensitivity coefficient.
    \item $v_b$ is the base speed of the robot.
\end{itemize}

\begin{algorithm}
\SetAlgoVlined


\While{power on}
{
    \eIf{No collision detected}
    {
        \eIf{On the LAP (blue trail)}
        {
            \eIf{On the LAP and SAP (cyan trail)}
            {
                \eIf{On the bifurcations of LAP and SRP (magenta trail)}
                {
                    Follow the LAP and SAP (cyan trail)\;
                    }
                    {
                    \eIf{On the bifurcation with LAP and SAP (cyan trail)}
                    {
                        Follow the LAP and SAP (cyan trail) with 70\% possibility\;
                        Follow the LAP (blue trail) with 30\% possibility\;
                        
                        }{Follow the LAP and SAP (cyan trail)\;
                    }
                } 
                }{Follow the LAP (blue trail)\;
            }
        
            }{Wandering\;
        }
        }{Avoiding\;
    }

 }
\caption{Ant foraging control strategy}
\label{algo:AntForaging}
\end{algorithm}

\subsection{Case II : Aggregation and Alarm Pheromone}
Case I showed the multiple pheromone involved in the same task with one single goal. In case II, we study how the multiple pheromones interact with each other and in sequence affect the agent's behaviour in different tasks. To implement this, we devise two tasks that the robots should do: first is to do the aggregation simulating the scenario of food carrying, another is to alert other individuals of the incoming predator to escape from being captured by the predator.  

\subsubsection{Pheromone Field Construction}
Two kinds of pheromones are designed inspired by the real insects, to guide the robots to do the aforementioned two different tasks: the aggregation pheromone (AGP) released by the leader robot to attract followers to gather together, the alarm pheromone (ALP) released by the robot which is close to the predator to repel others from the dangerous site where the predator is coming to. 

The pheromone in this study is modelled by \eqref{eq:PheroGaussian}, with the details of parameters' setting listed in Table \ref{tb:CS2_PP}. The injection of AGP is continuously updated according to the spontaneous position of the leader robot ($\vec{P_1}$). The injection of ALP can be refreshed and strengthened by any robots ($\vec{P_i},i
\in \{1,2,3,4\}$) whose distance with the predator agent is short enough.

\begin{table}
\renewcommand\arraystretch{2}         
\caption{Pheromone Parameter Setting in Case study II}
 \resizebox{0.48\textwidth}{!}{
\centering 
\begin{tabular}{cccccc}
\toprule
Type\hspace{-10pt}&Color\hspace{-5pt}&Model\hspace{-5pt}&Evaporation\hspace{-5pt}&Diffusion\hspace{-5pt}&Injection\hspace{-5pt}\\
\midrule
AGP\hspace{-10pt}&Green\hspace{-5pt}&\eqref{eq:PheroGaussian}\hspace{-5pt}&$e_\Phi = 10s$\hspace{-10pt}&\tabincell{c}{$\sigma_x = 40cm$ \\ $\sigma_y = 40cm$}&$(\mu_x,\mu_y) = \vec{P_{1}}$\hspace{-5pt}\\
ALP\hspace{-10pt}&Red\hspace{-5pt}&\eqref{eq:PheroGaussian}\hspace{-5pt}&$e_\Phi = 5s$\hspace{-5pt}&\tabincell{c}{$\sigma_x = 60cm$ \\ $\sigma_y = 60cm$}\hspace{-5pt}&\tabincell{c}{$(\mu_x,\mu_y)=\vec{P}_{i},$\\$i\in\{1,2,3,4\}$}\hspace{-5pt}\\
\bottomrule
\end{tabular}
}
\label{tb:CS2_PP}
\end{table}

\subsubsection{Robot Control}
The aim of the three followers is to aggregate around the leader without being captured by the predator. The predator freely wanders in the arena without special control. The leader and the follower strategy is described in Algorithm \ref{algo:Algo_preyLeader} and Algorithm \ref{algo:Algo_preyFollower} repectively.  And the method of controlling the robot to follow a direction is ruled by \eqref{eq:Case2Control}:

\begin{equation}
\left\{
\begin{array}{lr}
    R_l = v_b - p*(\theta_g - \theta_r) &\\
    R_r = v_b + p*(\theta_g - \theta_r)
\end{array}
\right.
    \label{eq:Case2Control}
\end{equation}

where:
\begin{itemize}
    \item $R_l$ and $R_r$ is the rotational speed of left wheel and right wheel respectively.
    \item $\theta_g$ and $\theta_r$ is the positive gradient direction of the aggregation and alarm pheromone respectively, which can be calculated by \eqref{Eq_Gradient}.
    \item $p$ is a sensitivity coefficient.
    \item $v_b$ is the base speed of the robot.
\end{itemize}

\begin{algorithm}
\SetAlgoVlined
\While{power on}
{
    \eIf{No collision detected}
    {
        \eIf{ALP (red)}
        {
            Calculate $(\theta_r)$ of ALP\;
            Move forward in this direction \eqref{eq:Case2Control}\;
            }{Wandering\;
        }
        }{Avoiding\;
    }

 }
\caption{ the leader control strategy}
\label{algo:Algo_preyLeader}
\end{algorithm}

\begin{algorithm}
\SetAlgoVlined


\While{power on}
{
    \eIf{No collision detected}
    {
        \eIf{ALG (red) $\|$ AGP (green)}
        {
            \eIf{ALG (red)}
            {
                Calculate $(\theta_r)$ of ALP (red)\;
                Move forward in this direction \eqref{eq:Case2Control}\;
                
                }{\If{AGP (green)}
                    {
                    Calculate $(\theta_g)$ of AGP (green)\;
                    Move forward in this direction \eqref{eq:Case2Control}\;
                    }
            }
            }{Wandering\;
        }
        }{Avoiding\;
    }
    
 }
\caption{the follower control strategy}
\label{algo:Algo_preyFollower}
\end{algorithm}

\section{Results and Discussion}
In this section we explain and discuss the results of the two case studies undertaken in the multiple pheromone communication system ColCOS$\Phi$ to verify this system's effectiveness and accuracy. Furthermore, how the multiple pheromones interacts with each other and in sequence affect the agents' behaviours will also be preliminarily discussed.
\subsection{Case I : Three Pheromones in Ant's Foraging}
\begin{figure}
    \centering
    \includegraphics[scale=0.45]{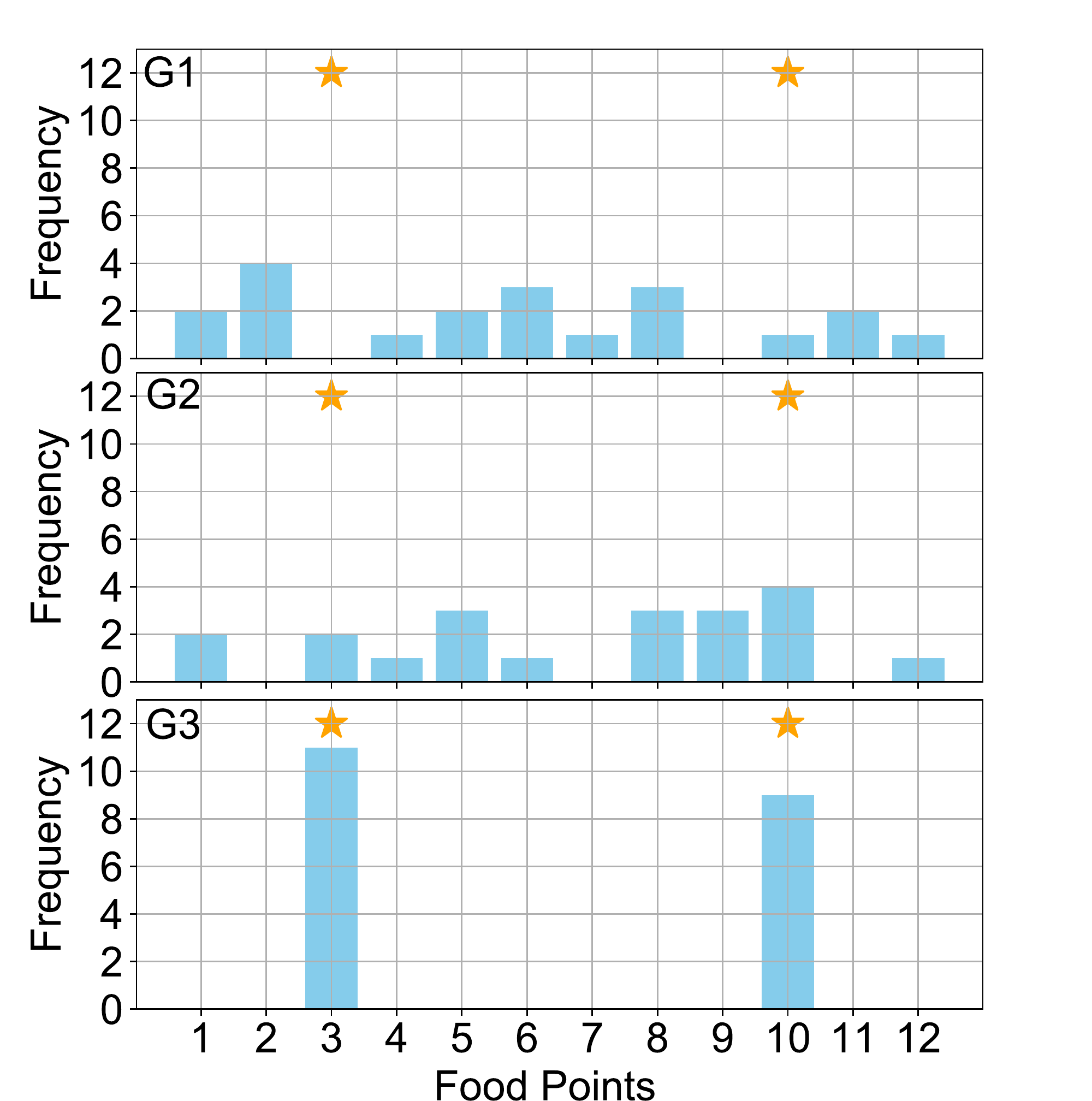}
    \caption{The arrival frequency distribution for each group of experiment. The orange stars indicate the current available food source, which in this case is $3$ and $10$. Note that sub-figures correspond to the pheromone maps depicted in Fig. \ref{fig:AntForageTrial}.}
    \label{fig:Fig_CS1_1}
\end{figure}

\begin{figure}
    \centering
    \includegraphics[scale=0.33]{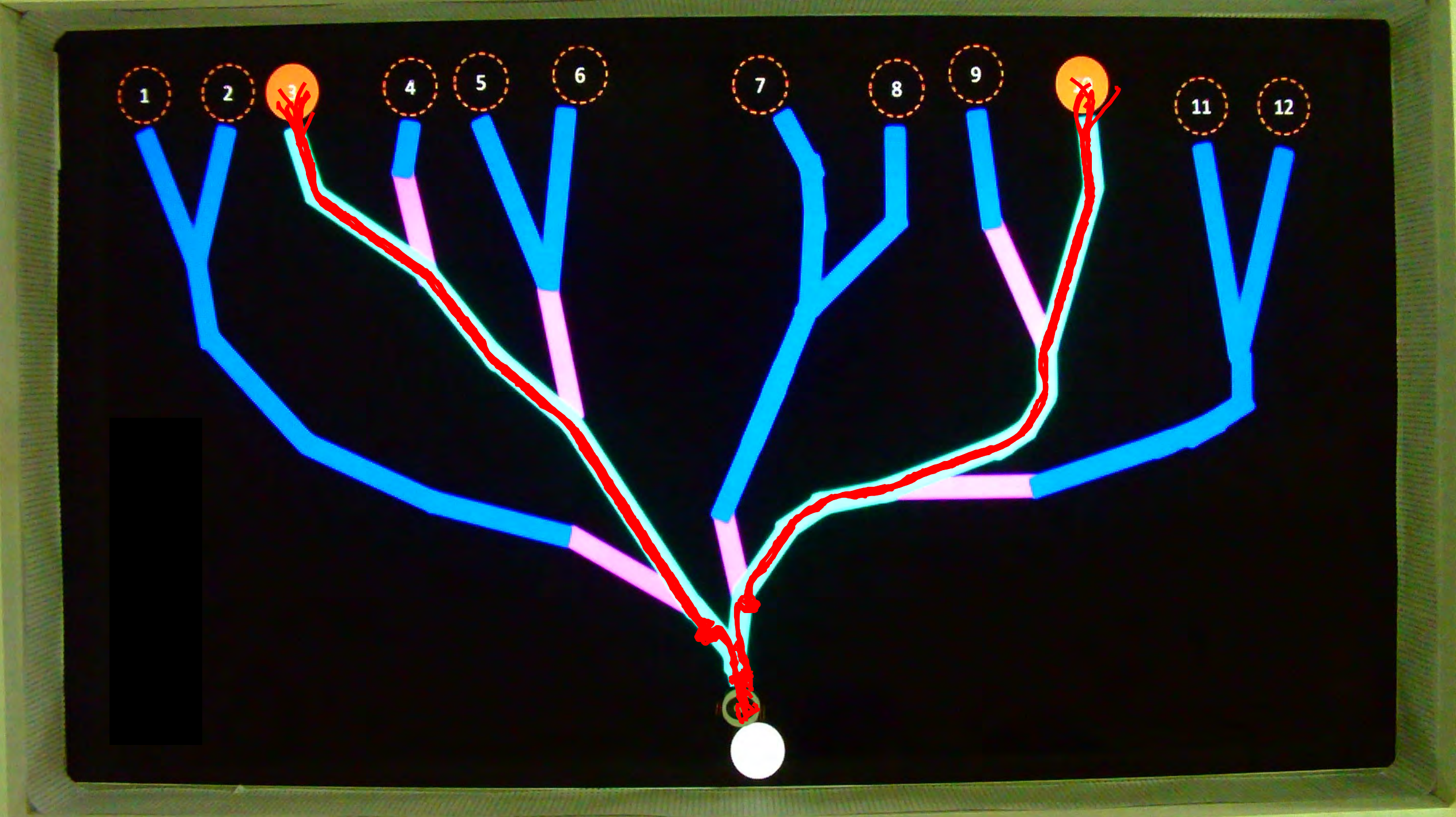}
    \caption{Foraging robot agent's trajectories ($20$ trials) in the case study I, the experiments of group G3. This background image is the screen shot of the video record from localisation system.}
    \label{fig:Fig_CS1_2}
\end{figure}

For case study I, we investigate the functions of these pheromones by devising 3 groups of experiments :
\begin{itemize}
    \item G1 : robot forages with only the LAP.
    \item G2 : robot forages with both the LAP and the SAP.
    \item G3 : robot forages with all the three pheromones.
\end{itemize}
Fig. \ref{fig:Fig_CS1_1} shows the foraging performance of one robot in three groups with constant current available food sources $3$ and $10$. The bar chart shows how many times the robot arrived at each food point in $N=20$ trails for each group. It is obvious that multiple pheromone significantly changes the distribution of the robot's arrival point and when all the three pheromones are presented, the robot will only arrive at the food points. Fig. \ref{fig:Fig_CS1_2} depicts the trajectories of the robot in G3 based on the output from the localisation system. The high overlap between the trajectories and the pheromone marked path in the background image shows the accuracy of the localisation system.

Experimental results of case study I have verified the effectiveness and flexibility of ColCOS$\Phi$, wherein different pheromones are independently represented and 
our robots can robustly distinguish different types of pheromones corresponding to proper behaviours.
Furthermore, this preliminary study of multiple pheromones involved in one task demonstrates how multiple pheromone can improve the efficiency, which may bring inspiration to biologists who are studying the multiple pheromones in social insects.



\subsection{Case II : Aggregation and Alarm Pheromone}
\begin{figure}
    \centering
    \includegraphics[scale=0.135]{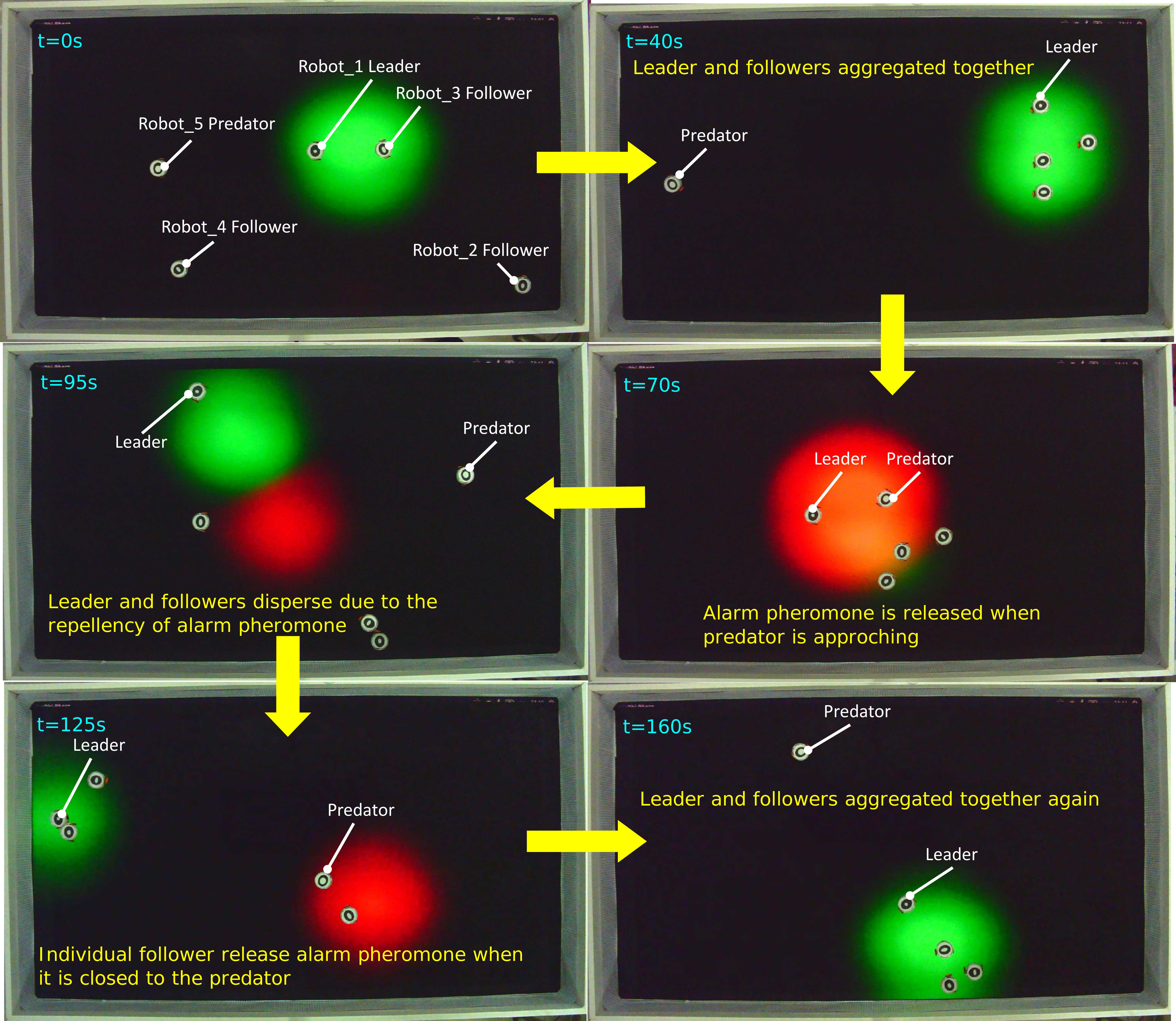}
    \caption{The snapshots of the experiment process of case study II. Yellow arrows represent the time flow. 
    }
    \label{fig:Fig_CS2_1}
\end{figure}

Fig. \ref{fig:Fig_CS2_1} shows the six typical snapshots of the experiment, from which we can see the swarm robots' aggregation behaviour guided by the attractive pheromone and the dispersion behaviour caused by the repellency of the alarm pheromone. 

\begin{figure}
    \centering
    \includegraphics[scale=0.25]{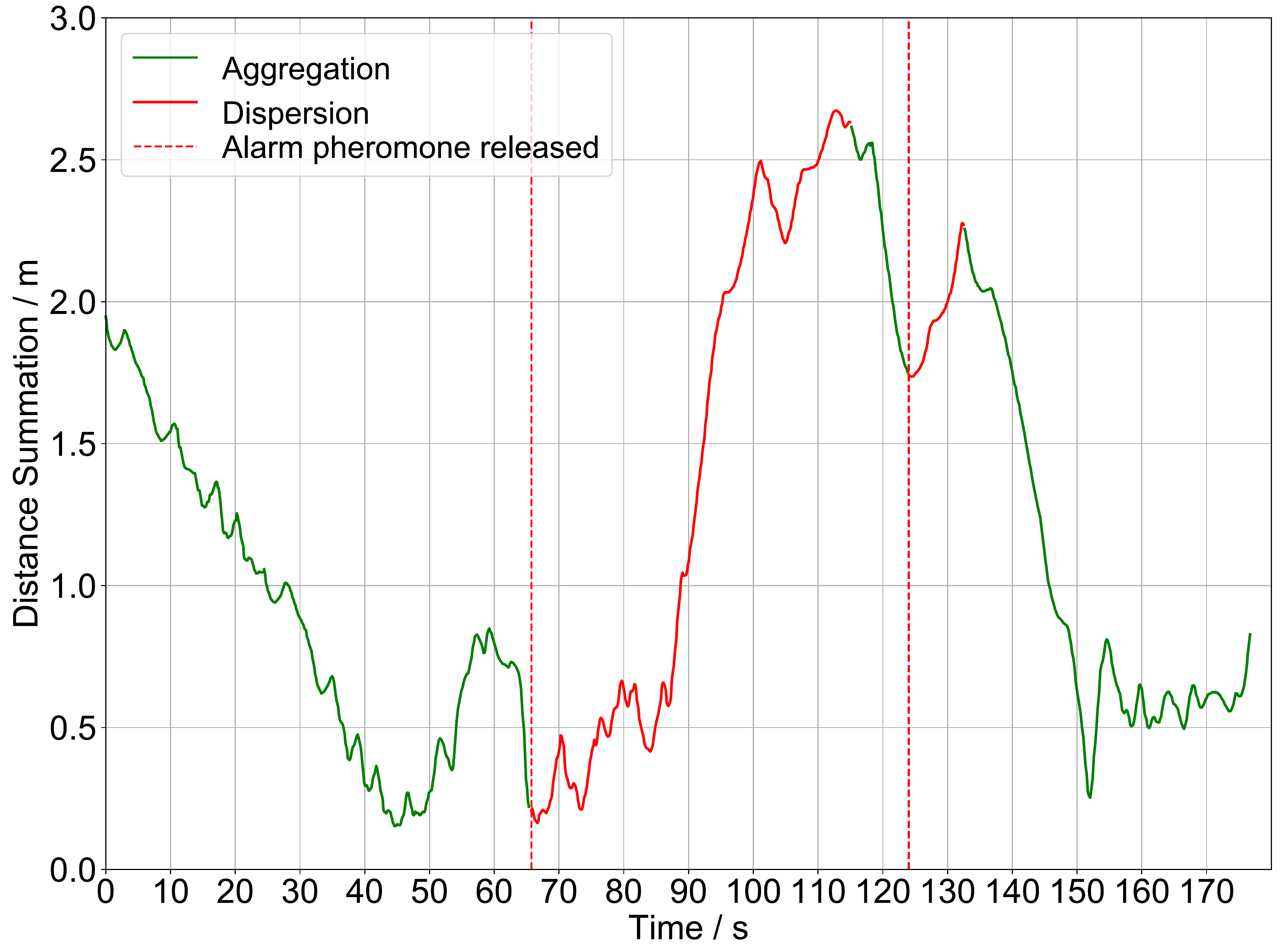}
    \caption{Summation of Euclidean distances between every follower and the leader overtime. Different swarm robot behaviours are indicated by different colours. The time stamp is in line with Fig.\ref{fig:Fig_CS2_1}}
    \label{fig:Fig_CS2_2}
\end{figure}

In order to evaluate how these two different pheromones affect swarm robots' behaviour, we calculate the summation of the Euclidean distances between the leader robot and the follower robots by \eqref{eq:SumDistance}:

\begin{equation}
    S(t) = \sum_{i=2}^{4}{\sqrt{(x_i(t)-x_1(t))^2 + (y_i(t)-y_1(t)^2}}
    \label{eq:SumDistance}
\end{equation}

where $(x_i(t),y_i(t)) = \vec{P}_i(t)$ is the $i^{th}$ follower's position at time $t$ and $(x_1(t),y_1(t)) = \vec{P}_1(t)$ is the leader's position at time $t$. The change of $S(t)$ during one experiment is showed in Fig. \ref{fig:Fig_CS2_2}, from which we can clearly see that attractive pheromone guided aggregation that leads to the decay of $S$ while the alarm pheromone caused dispersion resulting in the increase of $S$.

The results of case study II confirms the effectiveness and stability of our multiple pheromone system. In the dynamic scenario, the localisation system can track all the robot's locations and the virtual pheromone can be immediately released and updated on the screen which in sequence guides the swarm robots' behaviour, leading to complex collective behaviours like stigmergy. What's more, this study also suggests that complex and flexible swarm behaviours can be generated just via pheromone communications, which again verify the  meaning of multiple pheromone research.

Some limitations of this system also emerged during the experiments: such as the visual tracking algorithm is a little sensitive to the ambient light and there is a small delay of the pheromone releasing.

\section{CONCLUSIONS AND FUTURE WORK}
A compact and effective multiple pheromone communication system ColCOS$\Phi$ has been proposed.  The system, designed for the multiple pheromone research both in the swarm robotics and the social insects domains provides the possibility of studying the mechanisms of multiple pheromones with real robots in the natural and different dynamic environments for complex tasks. In addition, the newly redesigned micro-robot Colias IV with specific colour sensors on the bottom meets the requirements of sensing multiple pheromone cues. Cooperating with the overhead localisation system, it could easily implement the ideas within a researcher's mind and record the data of the experiment for later analysis. We can conclude from the two case studies that the system is robust, effective and flexible for multiple pheromone swarm robotics and social insects researches. 

For future work, firstly, the limitations of this system can be overcome by upgrading the hardware and software. Secondly, further studies and investigations can be undertaken based on the two case studies introduced in this paper, for example, the pheromone map in case study I can be dynamically created and modified by the robots instead of being generated artificially. Finally, more multiple pheromone scenarios can be devised and tested on this user-friendly system.

\bibliographystyle{IEEEtran} 
\bibliography{IEEEabrv,ref}

\end{document}